\documentclass[runningheads]{llncs}

\usepackage[mobile]{eccv}

\usepackage{eccvabbrv}

\usepackage{graphicx}
\usepackage{booktabs}

\usepackage[accsupp]{axessibility}  

\usepackage{multirow}
\usepackage{array}
\usepackage{graphbox}
\usepackage{color}
\usepackage{graphicx}
\usepackage{comment}
\usepackage{amsmath,amssymb} 
\usepackage{soul}
\usepackage{float}
\usepackage{pifont}
\usepackage{wrapfig}

\definecolor{linegreen}{RGB}{84, 130, 53}
\definecolor{lineorange}{RGB}{197, 90, 17}

\newcolumntype{?}[1]{!{\vrule width #1}}

\newcommand*{\ShowNotes}{}
\ifdefined\ShowNotes
  \newcommand{\colornote}[3]{{\color{#1}\bf{#2: #3}\normalfont}}
\else
  \newcommand{\colornote}[3]{}
\fi

\newcommand{\ctext}[3][RGB]{%
  \begingroup
  \definecolor{hlcolor}{#1}{#2}\sethlcolor{hlcolor}%
  \hl{#3}%
  \endgroup
}

\usepackage{hyperref}

\usepackage{orcidlink}

\begin{document}

\title{MedRAT: Unpaired Medical Report Generation  via Auxiliary Tasks} 


\author{Elad Hirsch \and
Gefen Dawidowicz \and
Ayellet Tal}

\authorrunning{E.~Hirsch et al.}

\institute{Technion – Israel Institute of Technology}

\maketitle

\begin{abstract}
Medical report generation from X-ray images is a challenging task, particularly in an unpaired setting where paired image-report data is unavailable for training. 
To address this challenge, we propose a novel model that leverages the available information in two distinct datasets, one comprising reports and the other consisting of images.
The core idea of our model revolves around the notion that combining auto-encoding report generation with multi-modal (report-image) alignment can offer a solution. However, the challenge persists regarding how to achieve this alignment when pair correspondence is absent. Our proposed solution involves the use of auxiliary tasks, particularly contrastive learning and classification, to position related images and reports in close proximity to each other.
This approach differs from previous methods that rely on pre-processing steps, such as using external information stored in a knowledge graph.
Our model, named {\em MedRAT,} surpasses previous state-of-the-art methods, demonstrating the feasibility of generating comprehensive medical reports without the need for paired data or external tools.
\end{abstract}

\section{Introduction}

The automated generation of medical reports has the potential to improve the efficiency and accuracy of patient information documentation, ultimately leading to better patient care and resource savings.
As a result, numerous studies have focused on this problem~\cite{chen-acl-2021-r2gencmn,chen-etal-2020-generating,jing2017automatic,li2019knowledge,wang2022cross}.
Although there are a few labeled image-report pair datasets available~\cite{demner2016preparing,johnson2019mimic}, they are relatively small compared to datasets for natural image~\cite{lin2014microsoft,sharma2018conceptual,thomee2016yfcc100m}. 
This is attributed to privacy concerns, limited access to high-quality data, and the complexity of medical data analysis and labeling, which require expert knowledge.

\begin{figure}[t]
  \begin{center}
    \includegraphics[width=1.0\linewidth]{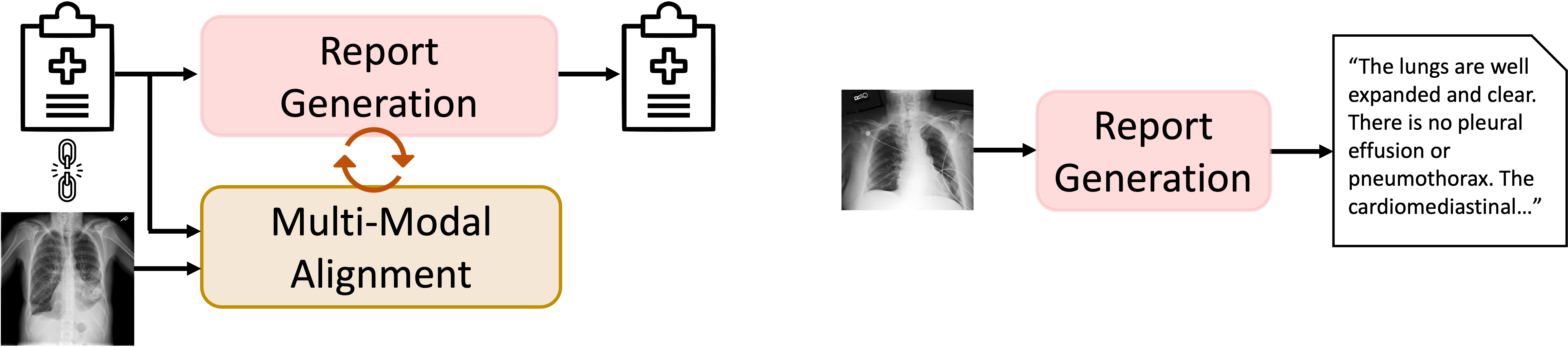}
  \end{center}
{ \small 
\hspace{0.59in}(a) Unpaired Training \hspace{1.2in} (b) Inference
}
\caption{{\bf Unpaired medical report generation.} 
 (a) We propose a model that addresses the challenge of unpaired images and reports by learning to generate reports from reports, and embedding related images and reports close together in the embedding space. Notably, our model achieves this without access to image-report pairs during training. (b) By learning both tasks simultaneously, our model is able to generate detailed reports for X-ray images during inference.
}
\label{fig:teaser}
\end{figure}

This paper focuses on the task of medical report generation in an unpaired setting, which addresses privacy issues and the shortage of paired data.
During training, two distinct datasets are available: one comprising chest X-ray images (e.g.,~\cite{irvin2019chexpert}) and the other containing chest X-ray reports (e.g.,~\cite{johnson2019mimic}). 
Notably, the images and reports are not paired with each other.
Each example in each dataset contains some domain information, typically related to the presence of specific pathologies. 
At inference, the model receives only an image and should generate a medical report.
The method of~\cite{liu2021auto} relies on a pre-constructed knowledge graph, whereas the recent method suggested in~\cite{hirsch2024medcycle} uses cycle-consistency and pseudo-reports.
Our approach eliminates the necessity for designing the graph or pseudo-reports.

Our method presents two main ideas to generate medical reports for images, without paired image-report data. 
Firstly, it learns how to generate the same report given as an input,
utilizing a language model and auto-encoding. 
Clearly, if we could only seamlessly replace the input report with an image during inference, our model would be able to generate reports from images.
Hence, the second idea is to approach the domain gap by learning to represent the reports and the images in a shared space that captures similarities and dissimilarities between them, i.e. multi-modal alignment.
Our novel architecture complements these goals by leveraging both global and local representations for the alignment and generation processes.
Differently from~\cite{hirsch2024medcycle,liu2021auto}, we do not rely on designing additional components (pseudo-reports or knowledge graphs), but instead solely use the provided data.
If performed effectively, it would generate high-quality reports (Figure~\ref{fig:teaser}(b)). 
We elaborate on the two sub-tasks below.

Although auto-encoding reports during training appears to be a straightforward approach, the limited diversity and size of the dataset can hinder generalization. 
To overcome this limitation, we propose a two-level approach.
At the global level, the report representation focuses on capturing the overall information, in particular the presence of pathologies. 
At the local level, the representation exposes the model to detailed information such as the location, size, and relation to other organs or devices~\cite{dawidowicz2023limitr}. 

To generate reports from images, rather than from reports, we aim to embed corresponding image and report pairs closely together, while pushing non-corresponding pairs apart. 
However, since the correspondence between images and reports is unknown, we propose three complementary techniques to achieve this embedding. 
First, we apply two auxiliary training tasks, multi-modal contrastive learning and multi-label classification, to embed semantically similar data points together. 
As our pair correspondence is unavailable, we use the existence or non-existence of certain common pathologies, as the only available global information, to position reports and images in a shared space. 
We show that both auxiliary tasks are necessary, due to the multi-label nature of the data.
Second, we propose a shared encoder-decoder architecture that is modality-independent. 
The shared encoder creates representations for both modalities in a shared space, and the shared decoder maps these representations to a report. 
Our encoder-decoder uses both global and local perspectives. 
Third, we construct a shared memory that is used for both modalities. 
This memory enables the model to record useful feature information and connections from past examples, shared by the two modalities, thus encoding domain knowledge.
Unlike in previous work, where hand-crafted knowledge was proposed, in our case, this knowledge is learned during end-to-end training.

Our method's effectiveness is demonstrated through improvements in both language metrics and clinical metrics.
For example, 
our method achieves  a $~10\%$ increase in clinical efficacy  (recall) on the test dataset from~\cite{johnson2019mimic} compared to~\cite{hirsch2024medcycle,liu2021auto}.
In this experiment, we use the images from~\cite{irvin2019chexpert} for training; this dataset lacks public reports, hence cannot be used by paired methods.

Hence, our paper makes the following contributions:
\begin{enumerate}
    \item 
    We introduce a novel method that addresses the challenge of unpaired X-ray report generation. 
    It learns to align multi-modal information via two auxiliary tasks---multi-label contrastive learning \& classification.
    The selection of these particular tasks is derived from the nature of our multi-label \& multi-modal domain.
    \item 
    We propose an architecture that leverages local and global representations. 
    While the global representation plays a crucial role in achieving multi-modal alignment, it alone is insufficient for generating the report. 
    Conversely, local representations are essential for detailed report generation, despite the absence of local alignments.
    \item
    Our method outperforms the SoTA results in unpaired chest X-ray report generation. 
   In addition, employing only a small amount of paired data can significantly improve the results further.
    It also significantly reduces the gap to models that use paired data. 
\end{enumerate}

\section{Related Work}
\label{sec:related}
There is limited research on generating medical reports through unpaired methods. 
Two closest related works are KGAE~\cite{liu2021auto} and the recently published MedCycle~\cite{hirsch2024medcycle}.
KGAE utilizes a pre-constructed knowledge graph to connect visual and textual modalities.
However, graph design requires domain expertise.
MedCycle, instead, relies on cycle-consistency and pseudo-reports to bridge the two modalities. 
Consequently, the design of the pseudo-reports might significantly affect the results.
In addition, mapping one modality to the other via cycle may still preserve hidden source information, just as a clue to map it back~\cite{nizan2020breaking}.
By training via auxiliary tasks in a shared space, we bypass these requirements.
We hereby elaborate on two related tasks.

\noindent
{\bf Paired medical report generation.}
Paired methods have access to pairs of images and corresponding reports.
 These models typically consist of an encoder and a decoder, with the encoder extracting visual features (usually by a CNN) and the decoder generating text. Some models use a hierarchical decoder composed of topic and word decoders~\cite{jing2017automatic,liu2019clinically,zhang2020radiology}, while others use transformers~\cite{chen-acl-2021-r2gencmn,chen-etal-2020-generating,huang2023kiut,li2023unify}.
 Knowledge graphs~\cite{li2019knowledge,liu2021exploring,zhang2020radiology} or memory blocks~\cite{chen-acl-2021-r2gencmn,chen-etal-2020-generating,wang2022cross,wang2022medical} are used to encode or learn domain knowledge and patterns. 
 Reinforcement learning techniques have been employed in some works to improve report consistency and accuracy~\cite{Gu_2024_WACV,miura-etal-2021-improving,nishino2020reinforcement}.

\noindent
{\bf Unpaired image captioning.}
Several studies have investigated captioning models that do not rely on image-caption pairs. 
These models are designed for the natural image domain, where there is sufficient auxiliary data or models to link vision and language. 
They employ efficient tools such as classifiers trained on large and diverse datasets, such as ImageNet, and external annotated sources~\cite{hendricks2016deep,venugopalan2017captioning}, or language-pivoting that utilizes large datasets in other languages~\cite{gu2018unpaired}. 
Object detectors that extract visual objects, concepts, and relationships have also improved the results of these models~\cite{feng2019unsupervised,gu2019unpaired,laina2019towards,liu2021exploring2,meng2022object}. 
Natural image captions mainly focus on describing the primary objects and their relationships.
Conversely, in the medical domain, the diagnosis is usually the main focus, which can be challenging to locate due to its size, distribution, relation to other organs, and limited available data.

\section{Method}
\label{sec:model}

 \begin{figure*}[t]
\centering
\includegraphics[width=1.0\linewidth]{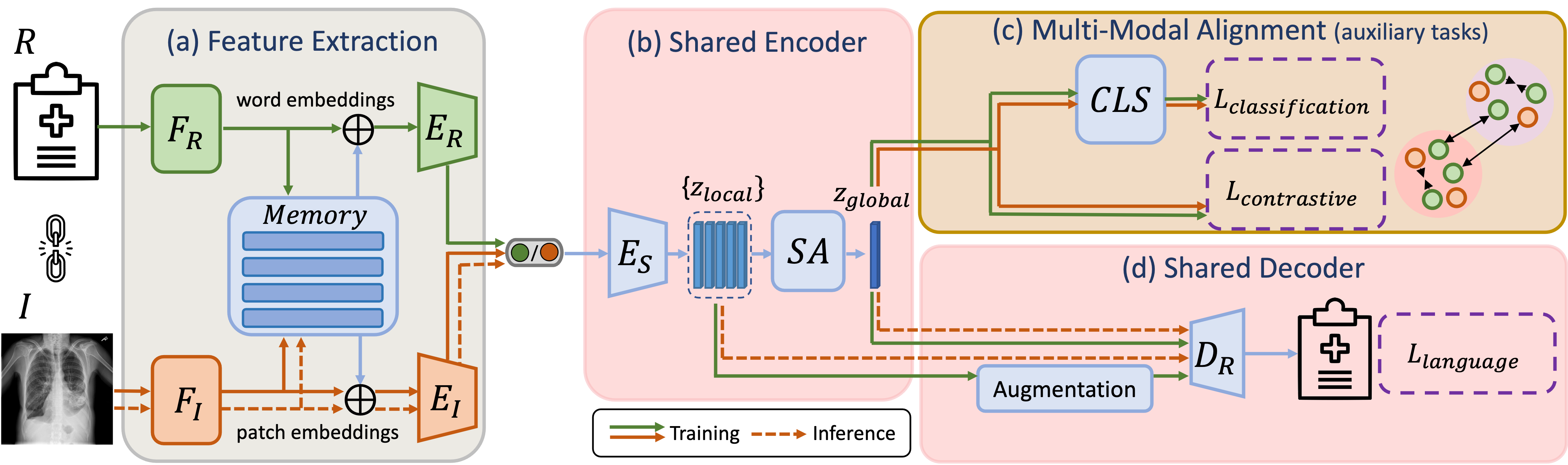}
\caption{{\bf Method.} 
(a) During the feature extraction stage, \textcolor{linegreen}{ reports} and \textcolor{lineorange}{images} pass through separate streams. 
Report words and image patches are encoded and combined with memory vectors that have been queried from the learned shared memory. 
(b)~The textual and visual embeddings are separately fed into a shared encoder, producing local representations in a shared space. 
These representations are then aggregated into a global representation using self-attention (SA). 
(c)~Multi-modal alignment is performed through auxiliary tasks---classification and contrastive learning---which pull closer relevant global representations or push them apart.
(d)~Simultaneously, local report representations are augmented during training, and the text decoder receives input from both the global and local representations, to produce the final report. 
During training, this is done with the report representations, while at inference, it uses the image representations. 
In this figure, the solid \textcolor{linegreen}{green} \& \textcolor{lineorange}{orange} lines indicate the training phase, while the dashed lines represent the inference phase.
}
\label{fig:method}
\end{figure*}

Our objective is to develop a model that can generate medical reports for X-ray images, without having access to image-report pairs during training. 
To achieve this, we propose to rely on two key ideas. 
The first is to train a language model to generate reports using an auto-encoder, based solely on reports.
Second, we learn to represent the reports and the X-ray images in a shared space that captures similarities and dissimilarities between them despite the lack of corresponding pairs. 
The challenge lies in achieving this joint representation without the use of paired data. Ultimately, this joint representation will be employed for generating a report from an image during inference.

\noindent
{\bf General structure.}
Our model architecture, depicted in Figure~\ref{fig:method}, integrates both key ideas.
It is composed of four blocks: 
(a)~feature extraction for either visual or textual data, 
(b)~an encoder that is shared between the two modalities, 
(c)~a multi-modal alignment module that learns two auxiliary tasks to encourage global corresponding representations to be similar (or dissimilar if they are not corresponding), and 
(d)~a decoder that generates reports from textual representations during training and from any representation during inference, particularly visual. 
It is worth noting that modules (a,b,d) implement both key ideas, while the multi-modal alignment module is essential solely for the second key idea. 

The joint embedding is implemented at every level of the network: The feature extraction module uses shared memory at the low-level (a), the local representations are embedded in a shared space at the intermediate level (b), and the auxiliary tasks that are modality-agnostic are utilized in the global level (c). 
We elaborate on each module below.

Overall, our novelty of the model can be outlined as follows:
(1) 
The approach we use to establish alignment between the global representations of images and reports is distinctive, involving a combination of two specific auxiliary tasks relevant to this domain.
(2) 
Our method for generating the report involves the utilization of both local and global representations, even in the absence of necessary local alignments, achieved through a self-attention mechanism.
(3) We address the challenge of generalization, given the susceptibility to overfitting in auto-encoding tasks, by incorporating dropout or noise into the input of the decoder.

\subsection{Feature extraction \& shared memory}
At the feature extraction stage for the image or report, any standard visual or textual extractor may be utilized.
We use a specific MLP, $F_I$\cite{he2016deep} or $F_R$\cite{bengio2000neural}, respectively. 
These features are passed to a shared memory matrix where they are enriched with additional information obtained through querying the memory. 
The querying process is agnostic to the source modality, allowing information to be shared between data points. 
The resulting enriched vectors are fed into their respective encoders, $E_I$ or $E_R$, to generate the visual and textual features.

\noindent
{\bf Shared memory.}
The memory serves two purposes: 
(1) to capture domain knowledge, such as priori information and relationships in the data, and 
(2) to bridge the gap between visual and textual features. 
While the former is a common use of memory in other domains~\cite{chen-acl-2021-r2gencmn,cornia2020meshed}, the latter is unique to our approach. 

Towards these goals, the memory used in our model is a trainable matrix of size $M \times d$, where $M$ is the number of vectors and $d$ is the feature dimension. 
During training and inference, the memory is queried based on the similarity of the given features and returns a subset of vectors that are aggregated with the input features and propagated forward in the network.
The memory is updated during backpropagation like any other parameter. 
Since the memory queries are based solely on feature similarity and are modality-agnostic, our model encourages image and report features to become more similar.

Specifically, a given feature vector $f$ is first projected using a trainable linear transformation $W_f$, to generate $f_p$. 
Similarly, the memory vectors ${m_1, \dots, m_M}$ are projected into a space of dimension $d$ using a trainable linear transformation $W_{m,in}$. 
Using cosine similarity, the top-K most similar projected memory vectors to $f_p$ are identified, resulting in the queried vectors ${m_{s_1}, \dots, m_{s_K}}$. 
Each queried vector $m_{s_i}$ is associated with its similarity $D_{s_i}$. 
To generate the memory response $r_f$, the queried vectors are projected using a trainable linear transformation $W_{m,out}$ and combined using a weighted sum based on their similarities $D_{s_1}, ..., D_{s_K}$. 
The overall expression for the memory response is as follows:
\begin{equation}
    r_f = \sum _{i=1} ^K \Big( \frac{exp(D_{s_i})}{\sum _{j=1} ^K exp(D_{s_j})} \Big) \cdot m_{s_i} \cdot W_{m,out} \, .
\end{equation}
Finally, the new feature vector is defined as $f_m = f + r_f$.
In backpropagation, the K-queried memory vectors are updated during the network update.

\subsection{Shared encoder-decoder}
\label{sec:shared-encoder-decoder}
Given feature vectors, of either a report or an image, the encoder maps them to a shared space where similarity can be measured. 
The decoder then transforms the representation into a sequence of words that make up the output report. 
During training, an auto-encoded report is produced as output, while during inference, a report that matches the input image representation is generated. 
To ensure the effectiveness of this approach, the representations of reports and images should be interchangeable.

\noindent
{\bf Shared encoder.}
The module receives  feature vectors and produces both local and global representations that serve as inputs to the decoder and to the alignment modules.
It is worth noting that from now on the trainable modules are agnostic to the data source.

The encoder $E_S$ maps embedding vectors from two different spaces into a single shared space; it is the target space for applying the auxiliary tasks discussed in Section~\ref{sec:multi-modal-alignment}.
The output of $E_S$ is a set of local representations, $\{z_{local}\}$, which correspond to either image patches or report words.

In order to create a global representation of an image or a report, the local representations need to be aggregated. 
However, not all local vectors are equally important, as some may correspond to insignificant features, hence simple averaging is not effective. 
Instead, we propose to learn the weights of the local representations  using self-attention~\cite{lin2017structured}, which considers the context of each vector within the image or report.
Furthermore, it can produce weights that are independent of the input size, which is important because the number of local representations can vary based on report length or the number of image patches.
A weighted average of the local vectors produces the global representation vector $z_{global}$, which is used as input for both the decoder and the alignment module. 

In terms of implementation, the encoder $E_S$ is composed of a series of multi-head attention layers (Transformer)~\cite{vaswani2017attention}, which have been proven effective in capturing relationships between various parts of the input.
For the self-attention component, let $Z_l$ be an $n \times d$ matrix ($n$ being the number of local representations) consisting of the vectors ${z_{local}}$ and $W_{SA_1}^{d \times d}$ and $W_{SA_2}^{1 \times d}$ be linear transformations.
The self-attention mechanism is implemented as follows:
\begin{gather}
A = softmax \big( W_{SA_2} \cdot \tanh (W_{SA_1} \cdot Z_l^T) \big), \\
z_{global} = A \cdot Z_l \, .
\end{gather} 

After softmax, $A$ is a weight vector of size $n$, with a sum of~$1$.
The global representation $z_{global}$ is a weighted average of the local representations, with the weights being the self-attention weights $A$.

\noindent
{\bf Shared decoder.}
The purpose of this module is to generate a report as output. 
Recall that during training, the module learns to map report representations to a sequence of words, while during inference, it generates a sequence of words based on the input image representations.
To ensure the accuracy of the reports, both the local (words/patches) and global (report/image) representations are used. 
The global representation captures the essence of the data, while the local representation captures the nuances and the details, including size, severity, relative position, etc.

The auto-encoding process has a potential problem of overfitting, which might hinder generalization. 
To overcome this issue, we propose to augment the local representations during training, which introduces variations and distortions that prevent the network from relying solely on memorizing patterns.
Two types of augmentations are proposed: (1) randomly masking the local representations using dropout, and (2) adding random noise to the local representations. 
We will demonstrate in Section~\ref{sec:ablation} that both augmentation types have a positive effect, yet one may be more suitable for a particular dataset.

Implementation-wise, $D_R$ consists of a sequence of multi-head attention layers (Transformer)~\cite{vaswani2017attention}.
$D_R$ iteratively predicts the next word of the sequence, based on the input and the past predicted words. 
In the final iteration, the entire report is generated, which is the output of the model.
As for the augmentation, when using dropout, each local vector has a probability $p$ of being masked out and not used by the decoder. 
Masking is random and independent of other vectors ($p=0.9$ in our experiments). 
The second method is adding additive noise, where a random vector sampled from a Gaussian distribution is added to every vector entry ($\sigma=5$ in our experiments). 

\subsection{Multi-modal alignment via auxiliary tasks}
\label{sec:multi-modal-alignment}
The aim of this module is to bring representations of corresponding images and reports closer while pushing apart non-matching ones, given their global representations. 
Since we lack matching pairs of images and reports during training, we propose to use the available domain information, which is the existence of certain pathologies. 
We incorporate this information by utilizing two auxiliary tasks that are not related to report generation, namely multi-label supervised contrastive learning and classification. 
Both tasks are derived from the global representations and applied in the shared space, making them oblivious to the modality source. 
We provide further details on each task below, and in Section~\ref{sec:ablation} we demonstrate the necessity of both tasks.

\noindent
{\bf Multi-modal contrastive learning.}
Contrastive learning is well-suited to our objective, which involves learning a feature representation that places semantically similar data points close together in the embedding space and separates those that are semantically dissimilar. 
To carry out this task, we must establish positive pairs within a batch. 
However, each batch contains $N$ images and $N$ unpaired reports, without correspondence information. 

We suggest to handle this issue by leveraging the available domain information to determine the positive pairs within a batch. 
We define a pair as positive if they share at least one pathology, which addresses the multi-label setting. Conversely, if a pair does not share any pathology, it is considered negative. 
Let $i$ denote the index of the example, $P(i)$ be the set of its positive pairs, $z_i$ be its global representation, and $\tau$ be a temperature parameter. 
The supervised contrastive learning loss is defined as follows~\cite{khosla2020supervised}:
\begin{equation}
\begin{aligned}
    L_{contrast} = \frac{1}{N} \sum _{i=0} ^{N-1} -\frac{1}{|P(i)|} \cdot \Biggl( \sum _{p \in P(i)} \log \frac{\exp (z_i \cdot z_p^T / \tau) }{\sum _{j=0,  \, j \neq i} ^{N-1} \exp (z_i \cdot z_j^T / \tau)} \Biggr).
    \label{eq:contrastive}
\end{aligned}
\end{equation}

For our case, the set $P(i)$ comprises both images and reports, allowing for addressing the multi-modal setting.
Moreover, we apply data augmentation within a single modality, which creates additional instances of the data points in a batch, where pairs of augmentations of the same data point are treated as positive pairs.
Specifically, we apply a sequence of random augmentations to images, such as crop, blur, affine transformations, and contrast change.
For reports, we randomly shuffle the inner-sentences.
In our implementation, we utilize two augmentations per data point, resulting in a replacement of $N$ in Equation~\ref{eq:contrastive} by $4N$.

It should be noted that the positive pairs defined using the  above method might not be entirely accurate, as an image may correspond to multiple reports and vice versa.
Furthermore, the number of shared pathologies is not taken into account.
The next task aims to address these inaccuracies.

\noindent
{\bf Multi-label classification.}
Our objective in the multi-label classification task is to predict the labels of each example, based on its global representation and the corresponding ground-truth.
Unlike contrastive learning, this task does not depend on pairing examples and is therefore immune to the inaccuracies arising from defining positive pairs.
Moreover, the classification task can compensate for partially-positive pairs, where only a subset of the pathologies match, which might lead to over-emphasizing their similarity in contrastive learning. 
By forcing the model to classify each example correctly, the partially-positive pairs are less likely to be pulled too close together.

The classification head in our implementation is composed of two linear layers with a ReLU activation function in between. 
The loss function used is the cross-entropy loss between the output of the classification head and the ground-truth labels. 
Let $y_i$ denote the ground-truth binary multi-label vector for the $i^{th}$ example (either report or image) and $x_i$ denote the prediction vector of the model for that example, after applying a sigmoid function. 
For a batch comprising $N$ images and $N$ reports, the loss is computed as follows:
\begin{equation}
\begin{aligned}
    L_{class} = - \frac{1}{2N} \sum _{i=0} ^{2N-1} \Bigl( y_i \cdot \log x_i^T + (\mathbf{1}-y_i) \cdot \log (\mathbf{1}-x_i)^T \Bigr).
    \label{eq:classification}
\end{aligned}
\end{equation}
\vspace{-0.3in}
%

\subsection{Losses}
\label{subsec:losses}
Our model optimizes a combination of three losses: language loss, contrastive loss and classification loss.
The language loss ($L_{lang}$) is responsible for generating coherent and accurate reports.
It is computed using the cross-entropy loss between the generated report and the ground-truth report.
The contrastive loss $L_{contrast}$ (Equation~\ref{eq:contrastive}) and the classification loss $L_{class}$ (Equation~\ref{eq:classification}) ensure that our model performs well during inference, when it is required to perform a different task from its training objective.
The overall loss is defined as:
\begin{equation}
\begin{aligned}
         L = \gamma_1 \cdot L_{lang} + \gamma _2  \cdot L_{contrast} + \gamma _3 \cdot L_{class}.
\end{aligned}
\end{equation}


\section{Experimental Results}
\label{sec:results}

{\bf Datasets.}
Our model was trained using images from the CheXpert dataset~\cite{irvin2019chexpert} and the training reports from the MIMIC-CXR dataset~\cite{johnson2019mimic}.
To evaluate the performance, we used test sets from two datasets: MIMIC-CXR and IU X-ray~\cite{demner2016preparing}.
During training, we utilized the images only for the auxiliary tasks, while matching reports were only used for evaluation.
Our experimental setup is similar to that of~\cite{liu2021auto}, with the exception that they also trained on reports from the IU X-ray dataset.
We followed the same report preprocessing steps as~\cite{chen-acl-2021-r2gencmn,liu2021auto}, where reports lacking a findings section were filtered out.
Importantly, there were no paired samples between the CheXpert and MIMIC-CXR or IU X-ray datasets.

{\em CheXpert}~\cite{irvin2019chexpert} is a large dataset of chest X-ray images, containing $224,316$ radiographs of $65,240$ patients, collected from Stanford Hospital.
Each image is labeled according to $14$ diagnosis classes
(multi-label).
While the labels are available to the public, the medical reports corresponding to the images are not.

{\em MIMIC-CXR}~\cite{johnson2019mimic} is a large dataset of chest radiographs, containing $377,110$ images, corresponding to $227,835$ reports performed at the Beth Israel Deaconess Medical Center.
Each radiograph is associated diagnostic labels as in~\cite{irvin2019chexpert}.
The dataset is split into $368,960$ images ($222,758$ reports) for training, $2,991$ images ($1,808$ reports) for validation, and $5,159$ images ($3,269$ reports) for testing.

{\em IU X-ray}~\cite{demner2016preparing} contains $7,470$ chest X-ray images associated with $3,955$ reports.
We use the same $70\%$-$10\%$-$20\%$ train-validation-test split, as set by~\cite{li2018hybrid}.

\begin{table*}[t]
\caption{\textbf{Quantitative evaluation, NLG metrics.}
Our results outperform those of~\cite{liu2021auto,hirsch2024medcycle} for most datasets and language generation metrics: BLEU (B), METEOR (M), ROUGE-L (R-L), RadGraph F1 (R-G), and BERTScore (B-S).
}
\centering

\setlength\tabcolsep{3pt}
\resizebox{1\linewidth}{!}{
\begin{tabular}{m{0.6in} |
>{\centering}m{0.3in} >{\centering}m{0.3in} >{\centering}m{0.3in} >{\centering}m{0.3in} >{\centering}m{0.3in} >{\centering}m{0.3in} | 
>{\centering}m{0.3in} >{\centering}m{0.3in} >{\centering}m{0.3in} >{\centering}m{0.3in} >{\centering}m{0.3in} >{\centering\arraybackslash}m{0.3in}}
\multirow{2}{*}{Method} & \multicolumn{6}{c|}{MIMIC-CXR~\cite{johnson2019mimic}} & \multicolumn{6}{c}{IU X-ray~\cite{demner2016preparing}} \\[0.01in]
& B-1 & B-4 & M & R-L & R-G & B-S & B-1 & B-4 & M & R-L & R-G & B-S \\[0.01in] \hline
\vspace{0.01in}
KGAE & 0.221 & 0.062 & 0.097 & 0.208 & - & - & 0.417 & 0.126 & 0.149 & 0.318 & - & - \\[0.01in]
MedCycle & 0.309 & 0.061 & 0.115 & 0.216 & 0.126 & 0.333 & \textbf{0.461} & \textbf{0.143} & \textbf{0.182} & 0.332 & 0.196 & 0.308 \\[0.01in]
MedRAT & \textbf{0.365} & \textbf{0.086} & \textbf{0.132}& \textbf{0.251} & \textbf{0.138} & \textbf{0.398}  & 0.455 & 0.129 & 0.171 & \textbf{0.349} & \textbf{0.204} & \textbf{0.427} \\
\end{tabular}
}

\label{table:quant_nlg}
\end{table*}

\begin{table}[t]
\caption{\textbf{Quantitative evaluation, CE metrics.}
Our results outperform those of~\cite{liu2021auto,hirsch2024medcycle} in terms of clinical efficacy (CE) metrics, on the MIMIC-CXR dataset. 
}
\centering
\setlength\tabcolsep{3pt}
\begin{tabular}{m{0.95in} |
>{\centering}m{0.5in} >{\centering}m{0.5in} >{\centering\arraybackslash}m{0.5in}}

Method & Precision & Recall & F1 \\[0.01in] \hline
\vspace{0.01in}
KGAE~\cite{liu2021auto} & 0.214 & 0.158  & 0.156 \\[0.01in]
MedCycle~\cite{hirsch2024medcycle} & 0.230 & 0.171  & 0.183 \\[0.01in]
MedRAT {\small (Ours)} & \textbf{0.285} & \textbf{0.265} & \textbf{0.227} \\
\end{tabular}
\label{table:quant_ce}
\vspace{-0.2in}
\end{table}

\noindent
{\bf Evaluation metrics.}
We assess our model's performance in two aspects: the quality of the generated language ({\em NLG}), and the quality of its clinical efficacy ({\em CE}).
To evaluate NLG, we utilize the common 
n-gram based BLEU~\cite{papineni-etal-2002-bleu}, METEOR~\cite{banerjee-lavie-2005-meteor}, and ROUGE-L~\cite{lin-2004-rouge} metrics, which measure the similarity between the generated reports and the ground-truth.
We also employ two recent evaluation metrics studied for this task in~\cite{yu2023evaluating}, RadGraph F1 \& BERTScore, which show high alignment with radiologists judgment.
To evaluate CE, we employ the CheXpert~\cite{irvin2019chexpert} model to assign $14$ diagnosis classes related to thoracic diseases and support devices, and we measure precision, recall, and F1 score with respect to the ground-truth labels.

\begin{table*}[t]
\caption{\textbf{Comparison to paired methods.}
The difference in performance between paired and unpaired methods narrows considerably with our MedRAT.
As expected, recent paired methods perform better (on the MIMIC-CXR dataset).
The column "Ratio" indicates the ratio between the specific method and the best paired one (underlined).
In addition, by utilizing only a small number of image-report pairs ($1\%$), our results improve further.
Notably, in several metrics (B-1, B-4 \& M) our few-shot results are comparable to those of some fully-paired methods (the ratio column is computed relative to the best method in each metric).
}
\centering
\small
\resizebox{1\linewidth}{!}{
\begin{tabular}{m{0.95in} |
>{\centering}m{0.5in} ?{0.4mm} >{\centering}m{0.3in} >{\centering}m{0.3in} | >{\centering}m{0.3in} >{\centering}m{0.3in} | >{\centering}m{0.3in} >{\centering}m{0.3in} | >{\centering}m{0.3in} >{\centering}m{0.3in} ?{0.4mm} >{\centering}m{0.33in} >{\centering\arraybackslash}m{0.3in}}
Method & \% Pairs & B-1 & Ratio & B-4 & Ratio & M & Ratio & R-L & Ratio & F1 & Ratio \\[0.01in] \hline
\vspace{0.01in}
R2Gen-CMN~\cite{chen-acl-2021-r2gencmn} & 100\% & 0.353 & 90\% & 0.106 & 85\% & 0.142 & 89\% & 0.278 & 94\% & 0.278 & 78\% \\
KGAE-Sup~\cite{liu2021auto} & 100\% & 0.369 & 94\% & 0.118 & 95\% & 0.153 & 96\% & \underline{0.295} & - & \underline{0.355} & - \\
COMG+RL~\cite{Gu_2024_WACV} & 100\% & 0.363 & 92\% & \underline{0.124} & - & 0.128 & 80\% & 0.290 & 98\% & 0.345 & 97\%\\ 
KiUT~\cite{huang2023kiut} & 100\% & \underline{0.393} & - & 0.113 & 91\% & \underline{0.160} & - & 0.285 & 97\% & 0.321 & 90\% \\
UAR~\cite{li2023unify} & 100\% & 0.363 & 92\% & 0.107 & 86\% & 0.157 & 98\% & 0.289 & 98\% & - & - \\ [0.01in] \hline
\hline
\vspace{0.01in}
KGAE~\cite{liu2021auto} & 0\% & 0.221 & 56\% & 0.062 & 50\% & 0.097 & 61\% & 0.208 & 71\% & 0.156 & 44\% \\
MedCycle~\cite{hirsch2024medcycle} & 0\% & 0.309 & 79\% & 0.061 & 49\% & 0.115 & 72\% & 0.216 & 73\% & 0.183 & 52\% \\
MedRAT {\small (Ours)} & 0\% & \underline{0.365} &\textbf{93\%} & \underline{0.086} & \textbf{69\%} & \underline{0.132} & \textbf{83\%} & \underline{0.251} & \textbf{85\%} & \underline{0.227} & \textbf{64\%} \\[0.01in] \hline \hline
MedRAT {\small (Ours)} & 1\% & 0.390 & \textbf{99\%} & 0.105 & \textbf{85\%} & 0.142 & \textbf{89\%} & 0.271 & \textbf{92\%} & 0.249 & \textbf{70\%} \\[0.01in]
\end{tabular}
}
\label{table:compare_supervised}
\vspace{-0.03in}
\end{table*}

 \begin{figure*}[t]
\centering
\begin{tabular}{>{\centering}m{0.02\linewidth} >{\centering}m{0.2\linewidth} m{0.36\linewidth} m{0.36\linewidth}}
     \rotatebox{90}{{\small Tags: \ctext[RGB]{252,241,192}{Edema}}} &
     \includegraphics[width=1\linewidth]{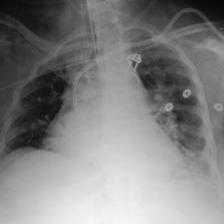} &
     {\footnotesize
    \ctext[RGB]{149,182,213}{Lung volumes are low} despite endotracheal intubation. 
    There is worsening engorgement of the mediastinal vasculature and \ctext[RGB]{255,214,165}{central pulmonary vascular congestion}. 
    There is also \ctext[RGB]{252,241,192}{new mild pulmonary edema} and bibasilar atelectasis. 
    There is \ctext[RGB]{192,185,191}{no pneumothorax} or pleural effusion. 
    The \ctext[RGB]{155,208,183}{endotracheal tube is in appropriate position approximately 5 cm above the carina}. 
    A hemodialysis catheter terminates in the cavoatrial junction.} &
    {\footnotesize
    There has been interval placement of an \ctext[RGB]{155,208,183}{endotracheal tube with its tip approximately 5 cm above the carina}. 
    An enteric tube is seen coursing below the level of the diaphragm inferior aspect not included on the image. 
    \ctext[RGB]{149,182,213}{Lung volumes are low}. 
    There is \ctext[RGB]{255,214,165}{prominence of the pulmonary vascular markings} suggestive of \ctext[RGB]{252,241,192}{mild pulmonary edema}. 
    There is possible small bilateral pleural effusions.
    The heart is mildly enlarged. 
    There is \ctext[RGB]{192,185,191}{no pneumothorax}.}\\
      & (a) Input image & \centering (b) Ground-truth report & \centering\arraybackslash (c) Our report \\
\end{tabular}
\vspace{-0.1in}
\caption{{\bf Qualitative evaluation.} 
Our model-generated report (c) contains similar information to the ground-truth report (b).
It describes the location of the \ctext[RGB]{155,208,183}{endotracheal tube tip} above the carina, the presence of \ctext[RGB]{252,241,192}{edema}, low \ctext[RGB]{149,182,213}{lung volumes}, and irregularities in the \ctext[RGB]{255,214,165}{pulmonary vascular}, while ruling out \ctext[RGB]{192,185,191}{pneumothorax} and suggesting only a possibility for small pleural effusion.
The report contains much more information than just the presence of edema (the information used for training) and uses similar phrases as the ground-truth report to describe the findings.
}
\label{fig:qualitative}
\end{figure*}
\noindent
{\bf Quantitative evaluation.}
Table~\ref{table:quant_nlg} presents a comparison between our method and previous works using NLG metrics. 
On the large and significant MIMIC-CXR dataset, our approach outperforms previous works in all metrics. 
On the IU X-ray dataset, our approach outperforms~\cite{liu2021auto} in all metrics and~\cite{hirsch2024medcycle} in some metrics, including RadGraph \& BERTScore, which align more closely with radiologist judgment than the other metrics.
We note that unlike~\cite{liu2021auto}, our method performs in a zero-shot fashion on this dataset, meaning that it was not exposed to any data from this dataset during training, neither images nor reports. 
%
Our reports not only resemble the ground-truth more closely, but are also more accurate and informative in extracting clinical information (pathologies), as shown in Table~\ref{table:quant_ce}. 
(As explained in~\cite{chen-acl-2021-r2gencmn}, these metrics do not suit IU X-ray dataset, due to its labeling schema, thus Table~\ref{table:quant_ce} focuses on MIMIC-CXR). 

Table~\ref{table:compare_supervised} presents a comparison between our results and those of recent paired and unpaired methods. 
Naturally, paired methods are expected to achieve better performance than unpaired methods.
However, our approach shows significant improvement and is even competitive by the BLEU-1 metric to paired methods, indicating a promising step towards closing the gap.
In addition, we demonstrate that utilizing a very small amount of paired data (1\% of the available paired data from~\cite{johnson2019mimic}) alongside our unpaired data leads to further improvements in the results.
This few-shot setting is valuable in scenarios where limited data can be paired by experts.
We compared our results on the new metrics BERTScore \& RadGraph to the paired methods of~\cite{Gu_2024_WACV, chen-acl-2021-r2gencmn} whose code is released.
For BERTScore: Ours $0.398$, Ours with few-shot $0.408$, \cite{Gu_2024_WACV} $0.397$, \cite{chen-acl-2021-r2gencmn} $0.413$; 
RadGraph: $0.138$, $0.164$, $0.171$ \& $0.180$, respectively.
These results indicate a good quality of our reports compared to the paired models.

\begin{figure*}[t]
\centering
\begin{tabular}{cccc}
     \includegraphics[width=0.2\linewidth]{figs/qualitative/2312/im} &
     \includegraphics[width=0.2\linewidth]{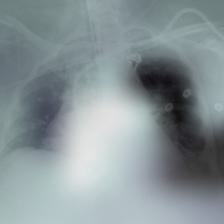} &
     \includegraphics[width=0.2\linewidth]{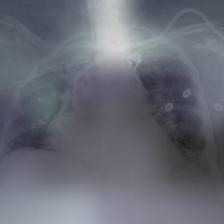} &
     \includegraphics[width=0.2\linewidth]{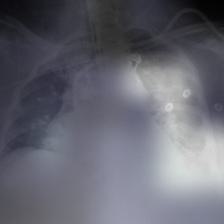} 
     \\
     (a) Input image & (b) Heart & (c) Tip & (d) Pleural \\
\end{tabular}
\caption{{\bf Attention visualization.} 
(a) presents the input and (b-d)  show examples of attention maps generated by our 
model, where bright values represent high attention. 
These maps demonstrate where the model is focusing when predicting specific words. 
For example, when predicting "heart" (b), the model's attention is on the central area around the heart; for "tip" (c), it concentrates around the trachea above the carina; and for "pleural" (d), it focuses on the left pleural cavity area (right side of the image).
Notably, this is achieved without training on patch-word alignment.
}
\label{fig:qualitative_attn}
\end{figure*}

\noindent
{\bf Qualitative evaluation.}
The similarities between our generated report and the ground-truth are depicted in Figure~\ref{fig:qualitative}. 
It is noteworthy that while the only global information associated  with the image is the presence of "Edema", our report includes additional details such as the type and severity of the edema.
Figure~\ref{fig:qualitative_attn} displays attention maps that highlight the decoder's focus while predicting specific words (blended with the input image). 
The maps show that the model identifies local connections between the image and the report when predicting words. 
For instance, the model focuses on the heart region while predicting the word "heart". 
Although the model was not trained on image-report pairs, the attention maps demonstrate its ability to establish relevant connections.

%

\section{Ablation Study}
\label{sec:ablation}

{\bf Contribution of the different components.}
%
The results in Table~\ref{table:ablation} indicate that the optimal performance is achieved by utilizing both the global and local representations during report generation. This could be attributed to the distinct functions of each representation; the global representation provides high-level semantic information that is shared across different reports and images, such as the existence of particular pathologies, whereas the local representations capture finer details, including descriptions of organs, abnormalities beyond those available globally, and spatial relationships.

\begin{table}[t]
\caption{\textbf{Ablation study.}
Each component contributes to the overall performance.
}
\centering
\small
\begin{tabular}{l|ccccc}
MedRAT & B-1 & B-4 & M & R-L & F1 \\[0.01in] \hline
\vspace{0.01in}
\hspace{2ex} {\small full} & \textbf{0.365} & \textbf{0.086} & \textbf{0.132}& \textbf{0.251} & \textbf{0.227} \\[0.01in] \hline
\hspace{2ex} decode w/o global & 0.316 & 0.068 & \textbf{0.132} & 0.242 & 0.224 \\[0.01in]
\hspace{2ex} decode w/o local & 0.162 & 0.020 & 0.082 & 0.180 & 0.092 \\[0.01in] \hline
\hspace{2ex} w/o contrastive & 0.271 & 0.046 & 0.131 & 0.215 & 0.196 \\[0.01in]
\hspace{2ex} w/o classification & 0.225 & 0.043 & 0.114 & 0.206 & 0.197 \\[0.01in] \hline
\hspace{2ex} w/o memory & 0.343 & 0.068 & 0.126 & 0.229 & 0.225 \\ 
\end{tabular}
\label{table:ablation}
\vspace{-0.2in}
\end{table}

The contrastive learning and classification tasks are the key of our approach, as auto-encoding reports without any connection to images is pointless. Table~\ref{table:ablation} indicates that combining both tasks results in better performance compared to using only one of them. 
This can be attributed to the fact that while supervised contrastive learning generalizes very well, it is designed for a single-label setup. 
In our multi-label setup, making a binary positive/negative decision for two samples is not straightforward.
As a result, only defining positive pairs as those that share at least one pathology might potentially bring partially-positive pairs too close together.
The classification task considers each sample separately and does not depend on setting the contrastive pairs. 
Thus, it enforces the embedding space to be informative enough to extract the ground-truth classes from each sample. 
However, employing only classification, thus treating each sample in isolation, might overlook the multi-modal relationships between them. 
This can result in an embedding space where two samples, originating from different modalities (i.e., an image and a report) but sharing the same labels, may end up being positioned too distantly from each other. 
It could lead to scenarios where samples from one modality form a distinct group, separated from the group of the other modality.
Thus, the two tasks---contrastive learning and classification---complement each other. 
Their combination ensures that samples from different modalities with shared labels are positioned in close proximity within the embedding space, fostering effective cross-modal associations, while
maintaining a necessary level of separation for samples that do not share all labels.

Finally, Table~\ref{table:ablation} also demonstrates that the inclusion of the memory component contributes to the improvement in performance.
This can be attributed to the memory's role in capturing domain knowledge and bridging the gap between visual and textual features.

\noindent
{\bf Augmentation.}
Recall that our model uses two types of augmentations, namely dropout and additive noise, for the local representations. Table~\ref{table:augment_ablation} demonstrates that the model's generalization is compromised without these augmentations. 
However, different datasets may benefit from different augmentations. 
Specifically, we observe that on MIMIC-CXR, dropout yields the best results, whereas on IU X-ray, Gaussian additive noise is more effective. 
These differences in performance may be attributed to variations in the statistics of the training and test datasets. 
For instance, the training image dataset, CheXpert, is noticeably different from IU X-ray in terms of image sharpness, contrast, and other factors, which makes noise a helpful augmentation. 
On the other hand, CheXpert and MIMIC-CXR are more similar in appearance, as evidenced by their similar pixel value standard deviations (between $73$-$76$) compared to IU X-ray (which has a standard deviation of $60$).

\begin{table}[t]
\caption{\textbf{Local augmentation methods.}
Augmenting the input of the text decoder improves the results.
Different augmentations may suit different target datasets.
}
\centering
\small
\begin{tabular}{c|ccc|ccc}
\multirow{2}{*}{\shortstack[c]{Augmentation \\ Method}} & \multicolumn{3}{c|}{MIMIC-CXR~\cite{johnson2019mimic}} & \multicolumn{3}{c}{IU X-ray~\cite{demner2016preparing}} \\[0.01in]
& B-1 & B-4 & M & B-1 & B-4 & M \\[0.01in] \hline
\vspace{0.01in}
None & 0.295 & 0.056 & 0.117 & 0.299 & 0.070 & 0.131 \\
Dropout & \textbf{0.365} & \textbf{0.086} & \textbf{0.132} & 0.333 & 0.075 & 0.136 \\
Noise & 0.342 & 0.074 & 0.126 & \textbf{0.455} & \textbf{0.129}& \textbf{0.171}\\
\end{tabular}
\vspace{-0.1in}
\label{table:augment_ablation}
\end{table}

\noindent
{\bf Limitations.}
First, our method generates a report based only on a given image. However, medical reports often include references to previous examinations (e.g. "in comparison," "improvement"). 
Our model is not capable of making such inferences, despite the significance of such information.
Second, 
our method requires common labels of thoracic pathologies; these however are available for all available large image and report datasets such as~\cite{irvin2019chexpert} and~\cite{johnson2019mimic}.

\noindent
{\bf Potential Impact.}
Our method utilizes unpaired data, mitigating the risk of patient privacy leaks.
In turn, the opportunities in employing unpaired data may encourage data owners to release more data.
However, it's important to acknowledge potential risks inherent in automated systems, as they might be susceptible to errors.
Given the significant impact of misdiagnoses, we advocate for these systems to complement radiologists rather than replace them.

\section{Conclusions}
This paper introduces a novel method for generating X-ray reports in an unpaired setting, without paired images and reports available during training. 
The proposed approach involves auto-encoding reports while simultaneously learning representations through two auxiliary tasks, namely contrastive learning and classification.
These tasks leverage available domain knowledge for each modality. 
We demonstrate the effectiveness of our approach on two different datasets.
It outperforms the state-of-the-art unpaired methods for chest X-ray report generation and significantly narrows the performance gap with paired methods.

\noindent
{\bf Acknowledgements.}
We gratefully acknowledge the support of the Israel Science Foundation 2329/22.

\clearpage

%
%

\bibliographystyle{splncs04}
\bibliography{main}
\end{document}